\documentclass[conference]{IEEEtran}
\usepackage{cite}
\usepackage{graphicx}
\graphicspath{ {images/} }
\IEEEoverridecommandlockouts
\usepackage{cite}
\usepackage{amsmath,amssymb,amsfonts,bbm}
\usepackage{subfigure,bm,balance}
\usepackage[ruled,vlined]{algorithm2e}
\usepackage{graphicx}
\usepackage{textcomp}
\usepackage{xcolor}

\begin{document}


\title{ERATTA: Extreme RAG for enterprise-Table To Answers with Large Language Models}
\author{\IEEEauthorblockN{Sohini Roychowdhury\IEEEauthorrefmark{1},
Marko Krema\IEEEauthorrefmark{1},
Anvar Mahammad\IEEEauthorrefmark{1},
Brian Moore\IEEEauthorrefmark{1}, 
Arijit Mukherjee \IEEEauthorrefmark{4},
Punit Prakashchandra \IEEEauthorrefmark{4}}
\IEEEauthorblockA{\IEEEauthorrefmark{1}Corporate Data and Analytics Office (CDAO), Accenture, USA}\IEEEauthorrefmark{4}CDAO, Accenture, India} \vspace{-0.5cm}
\maketitle
\begin{abstract}
Large language models (LLMs) with retrieval augmented-generation (RAG) have been the optimal choice for scalable generative AI solutions in the recent past. Although RAG implemented with AI agents (agentic-RAG) has been recently popularized, its suffers from unstable cost and unreliable performances for Enterprise-level data-practices. Most existing use-cases that incorporate RAG with LLMs have been either generic or extremely domain specific, thereby questioning the scalability and generalizability of RAG-LLM approaches. In this work, we propose a unique LLM-based system where multiple LLMs can be invoked to enable data authentication, user-query routing, data-retrieval and custom prompting for question-answering capabilities from Enterprise-level data tables on sustainability. The source tables here are highly fluctuating and large in size storing carbon footprint, energy and water usage at buildings in regional levels globally and the proposed framework enables structured responses in under 10 seconds per query. Additionally, we propose a five metric scoring module that detects and reports hallucinations in the LLM responses. Our proposed system and scoring metrics achieve $>90\%$ confidence scores across hundreds of user queries in the sustainability, financial health and social media domains. Extensions to the proposed extreme RAG architectures can enable heterogeneous source querying using LLMs.
\end{abstract}

\section{Introduction}\label{intro}
Generative AI solutions using Large Language Models (LLMs) have been extremely popular since the launch of Open AI's chatGPT-3 in late 2022 \cite{openai}. Most LLM solutions today can typically be categorized as Gen-AI applications, such as a chatbot-like experience with a virtual assistant for code, language and content generation \cite{bigdata}; or Gen-AI agents that are capable of automated orchestration to perform specific tasks such as booking tickets, writing blogs, articles and generating software automatically \cite{genaiagent1}. Most such solutions implement the retrieval augmented-generation (RAG) approach, wherein the most relevant data sources are first identified followed by the knowledge-graph (KG) method to isolate only relevant data entities that are collated with natural language instructions and answering guidelines and sent to the LLM to receive the required responses \cite{RAG1}. Although agentic-RAG approaches have been found to enhance the ``knowledge-quality'' of retrieved content instead of relying on LLM memory, these approaches can be costly and time-intensive owing to the web-search and KG method. Thus, for \textit{Enterprise-level} data products involving large numbers of tabular rows and columns, agentic-RAG can prove impractical for question-answering for a large number of users/groups. In this paper, we present a novel system architecture that maximizes the RAG impact through multiple LLM calls for tasks such as user authentication, query routing and natural language question-answering while ensuring security/access requirements and accuracy standards by flagging hallucinations for critical domains such as sustainability.

So far, towards use-cases that require question-answering from large volumes of tabular data at \textit{Enterprise} level, language-based RAG approaches have been accurate as presented in \cite{bigdata}. The proposed framework enhances system scalability by compartmentalizing the RAG process for specific tasks (extreme RAG) and by relying on automated code generation as an intermediate step for tables-to-answers use-cases. The various RAG tasks are as follows: 1) user-access authentication for tabular data based on minimal-user-profile (MUP) data, 2) user-query understanding and routing based on intention 3) SQL code generation from natural language to retrieve specific sub-tabular data (also knows as seq2sql \cite{seq2sql}) 4) response generation using the retrieved sub-tabular context and natural language. The proposed extreme RAG and hallucination control metrics presented here are scalable and efficient for knowledge retrieval with lower maintenance and operating costs when compared to the existing works \cite{bigdata} \cite{RAG1}. 
\begin{figure*}[ht]
    \centering
    \includegraphics[width=6.0in]{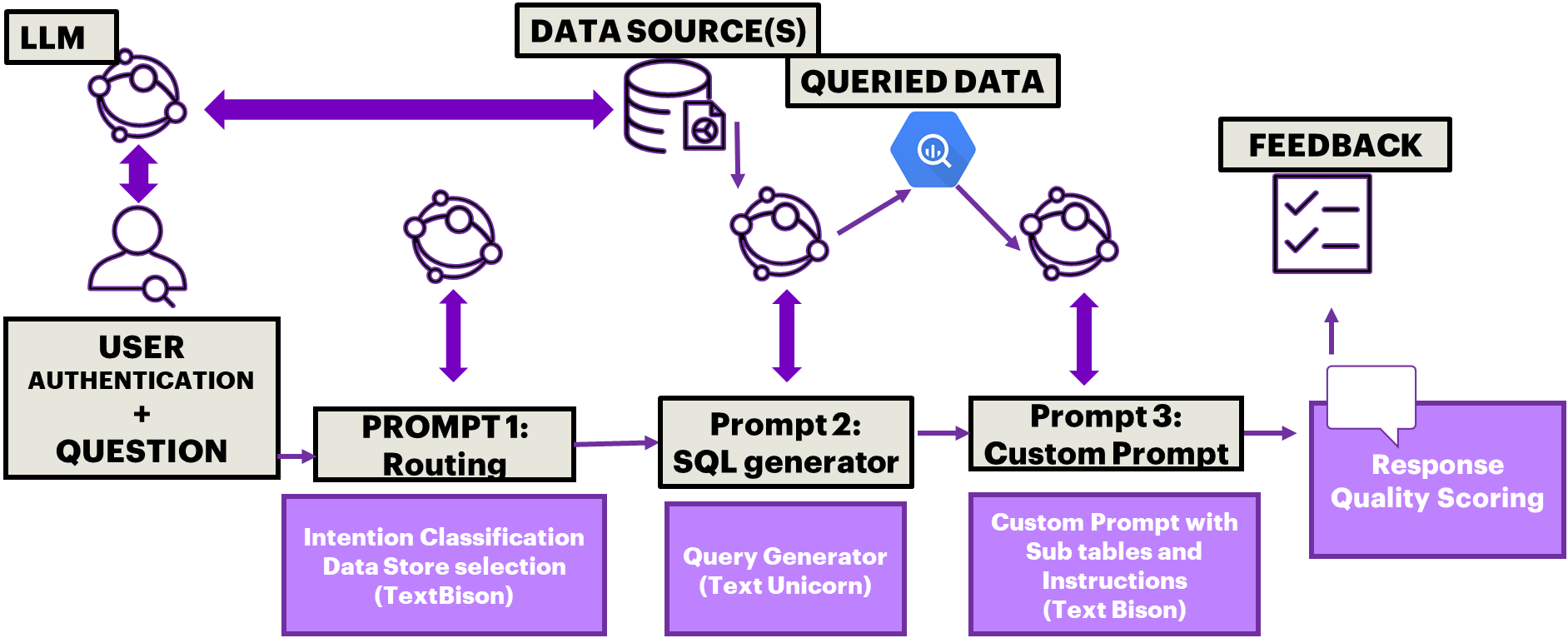}
    \caption{The proposed extreme-RAG system architecture that invokes 3-4 LLM prompts per user query to authenticate, route, extract smaller tabular data and fetch the required response from fast-varying and large tabular sustainability data sources.}
    \label{fig:sys} \vspace{-0.3cm}
\end{figure*}
The process of generating code (SQL, Python, C/C++) using LLMs to query tabular-data has several advantages. First, LLMs can accept data in tabular format, which allows better manageability for \textit{relevant data} search to each user query when compared to semantic retrieval on textual data from vector databases as presented in the work \cite{bigdata}. Structured data tables eliminate the need to manage a complex retrieval engine and results ranking system. Second, passing data in smaller sub-tables to LLMs enables scalability on new data tables without the need for the an offline table to text generation and vector embedding process in \cite{bigdata}. For tabular data sets that have high refresh rates such as stock market prices, inflation indices, carbon footprint etc., passing the relevant sub-tables ensures most accurate data-to-answers. Third, tabular data is structured and concise, which minimizes the scope for data misrepresentation and hallucinations. It is noteworthy that LLMs have been found to be more accurate and stable for text-to-code generation than for text-to-text generation in terms of \textit{fake responses} or \textit{hallucinations} \cite{openai}. Thus, our novel extreme-RAG system shown in Fig. \ref{fig:sys} incorporates text-to-SQL code generation followed by passing shorter data tables to LLM prompts to enable a variety of table-to-answers that are tested on \textit{Sustainability domain} datasets.

This paper makes three key contributions. First, we present a serial framework that implements RAG at multiple levels to complete specific tasks per step to separate the user-query understanding, data access and response generation processes. Second, we present non-LLM hallucination detection metrics that evaluate the accuracy and reliability of each query response. Third, we present extensions to the extreme-RAG framework towards answering predictive and prescriptive user-queries from Enterprise-tabular data. We compare the performance of the extreme-RAG with agentic-RAG systems and observe higher reliability and accuracy for the extreme RAG approach. This is primarily due to the intermediate stage of (Prompt 2 in Fig. \ref{fig:sys}) where SQL queries are automatically generated to retrieve the most relevant data from the enterprise tables. By focusing on smaller, sub-tables with labeled information, our system capitalizes on the LLM's strengths in handling concise and well-defined datasets. 
\section{Related Work}
Generating responses from tabular data sources has been an area of focus for the past few decades. Early works such as \cite{table2answer} relied on the structure of tabular data in xml/html documents to parse and extract the required answers. Prior to the advent of LLMs, some major efforts in this domain included the development of algorithms known as `seq2sql' models that can accept a sequence of input words and convert it into a corresponding SQL query as shown in Table \ref{tab:ex}.
\begin{table*}[ht] 
\caption{A summary of SEQ2SQL algorithms prior to LLMs for tables-to-answers use-cases and their capabilities.}
\scalebox{0.88}
    {
\begin{tabular}{|c|c|l|}
\hline
 Model Name&Year& Core Capabilities\\ \hline
 Bidirectional Attention &June, 2018&- Three sub-modules were designed with deep learning models for inference.\\
 for SQL Generation &&- Bi-directional attention mechanisms and character embeddings were implemented with convolutional neural networks (CNNs).\\
\cite{bisql}&&- Experimental evaluations were presented for the Wiki SQL dataset (87,726 hand annotated text and SQL).\\ \hline
Tranx&2018&- A transition-based neural semantic parser that mapped natural language (NL) utterances into formal meaning representations.\\
\cite{tranx}&&- Highly generalizable system that can be applied to new representations by just writing a new abstract syntax description.\\ 
&&- Experimental verification performed for 4 semantic parsing and code generation tasks.\\ \hline
STAMP+RL&2018&- For typical seq2SQL models, inaccuracies occur due to mismatch between question words and table contents.\\
\cite{stamp}&&- Quality of generated SQL query was improved through content replication by column name, cells or SQL keywords.\\
&&- Generation of `WHERE' clause was further improved by using column-cell relations. \\ \hline
PT-MAML&2018&- Adapted meta learning to solve the situation where the number of conditions in a text-to-SQL query vary significantly.\\
\cite{maml}&&- This method decoded a text sequence guided by a fixed syntax pattern by tagging for schema and constant values.\\ \hline
\end{tabular}\label{tab:ex}
}\vspace{-0.3cm}
\end{table*}
In this work, we extend the state-of-the-art by breaking down the tasks of data fetching from an input sequence of words into three stages, 1) query routing, 2) multiple-SQL generation, 3) SQL and text combination to yield accurate solutions to complex, multi-tabular queries from enterprise data products.
\section{Data and Methods}
 For our experiments, we utilize enterprise-scale \text{sustainability}-specific datasets containing information regarding carbon footprint, water/electricity consumption and renewable energy usage across countries and continents measured at daily, monthly and yearly levels. The dataset comprises of 7 tables of sizes 50 MB to 1.2 GB each with over 1000 rows and 50 columns of data in each table. Based on these tables, the user-queries can be categorized into the following 7 major categories and their combinations as shown in Table \ref{e2e}.
\begin{table}[ht]
\caption{Categories of Questions routed by Prompt 1}
\scalebox{0.8}
    {
\begin{tabular}{|l|l|l|}
\hline
{\bf Intention}&{\bf Type}&{\bf Example user-queries}\\
\hline
0&Percent&What \% of our offices are at 100\% renewable electricity?\\\hline
1&Change&What is the annual reduction of emissions globally?\\\hline
2&Rank&Which country has the highest Emissions type 1 emissions?\\\hline
3&Level&What is scope 1 emission levels for offices in Argentina?\\\hline
4&Rank&Which city had the highest water consumption for Dec 2022?\\\hline
5&Multi&Which countries reduced scope 3 emissions consistently in \\ 
&& the last 2 years and increased renewable electricity?\\\hline
6&FAQ&What is included in business travel?\\\hline
\end{tabular}\label{e2e}
\vspace{-0.3cm}
}
\end{table}
The descriptions for the extreme-RAG system components are presented below.
\subsection{Extreme-RAG System Components}
\subsubsection{Authentication RAG}
This process is a scalable extension to an otherwise rule-based look-up that enables/maps tables to users based on the user-access restrictions. As shown in Fig. \ref{fig:sys}, upon each user login, there is a need to enable a database lookup to retain access only to the tables that the user can retrieve from and to withhold access to all remaining tables. For example, a user from `North America' can only have access to power-consumption information from North America. So only the tables with user-access are pre-loaded to the Postgres DB or BigQuery for to set up a question-answering session while all remaining geo-location tables are skipped. Next, a comprehensive table-access profile (or MUP) needs to be generated to accommodate super-users who may need to access multiple tables across continents. For instance, a user specializing in renewable energy may require access to specific data tables across continents. For authenticating access for hundreds and thousands of users on thousands of data tables with varying schema, a custom prompt that outputs the minimal access constraints (or MUP) in json/xml formats is the first step for a chat session set-up.

\subsubsection{Prompt 1, Routing a User-query}
Once authenticated, each user-query needs to be routed for its intention and appropriate data tables. This query understanding and routing is an essential step since the instructions to respond to the different categories of questions in Table \ref{tab:ex} vary significantly. Query extension and query rewriting is then performed by converting the user-query into its high-dimensional embedded format and matching it to at-most five prior query samples in the vector space. The output of this prompt 1 is a comprehensive list of data sources relevant to the user-query.

This routing prompt is trained on a specific template to identify table relevance. The result is a list of table names in Python list format, which can be easily integrated with other data-processing tasks. Although this can be envisioned as a standard classification task as in \cite{bigdata}, LLMs are found to be more generalizable and scalable across a wide range of use queries and source tables with minimal training than standard supervised learning models. 

Following the identification of \textit{relevant} data sources, the proposed system proceeds to retrieve a list of data source configurations for each identified source. These data source configurations represent tabular metadata \cite{maml} and are important to isolate the appropriate SQL queries in the next step.
\subsubsection{Prompt 2, Data Retrieval}
Once the appropriate data tables are mapped to a user query, the data retrieval prompt converts standard language to SQL code. This prompt takes in 3 inputs: 1) the rewritten and extended sub-queries, 2) a list of data source configurations (metadata) obtained by semantic matching of the queries with prototype questions, and 3) a sample question and its answer. Utilizing these inputs, Prompt 2 generates a complex, nested, SQL query. A query runner process runs this SQL query against the data sources specified by the previous step to fetch the required information in tabular form. Only the \textit{relevant} rows and fields from the pre-loaded tables are fetched and loaded into a specific BigQuery Table. 

\begin{figure}[ht!]
    \centering
    \includegraphics[width=3.2in]{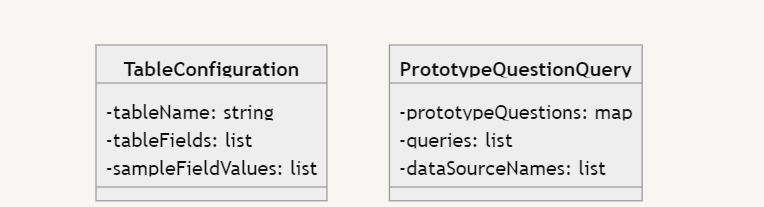}
    \caption{Examples of class definitions to extract tabular metadata and enable scaling prompt 2 across source tables and user query types.}
    \label{fig:p2}
\end{figure}

The metadata corresponding to the mapped tables, can be expressed as ``Table Configurations'' and the example questions can be specified as ``Prototype Question configurations'' as shown in Fig. \ref{fig:p2}. Here, \textit{Table Configurations} represents an encapsulation (class) with properties for the table name, a list of relevant table fields, and a list of sample field values. \textit{Prototype Question Query} similarly is another class that maps prior user-queries to data source names. The primary advantage of using these data structures in Prompt 2 is that it eliminates the need for code changes whenever the data domains change. For each new data source, the Table Configuration List is automatically updated, and with each new user-query category, the Prototype Question Query list is updated. Thus, these data structures enable generalizability and scalability. 
\subsubsection{Prompt 3, Answer Retrieval}
\begin{figure*}[ht!]
    \centering
    \includegraphics[width=6.2in]{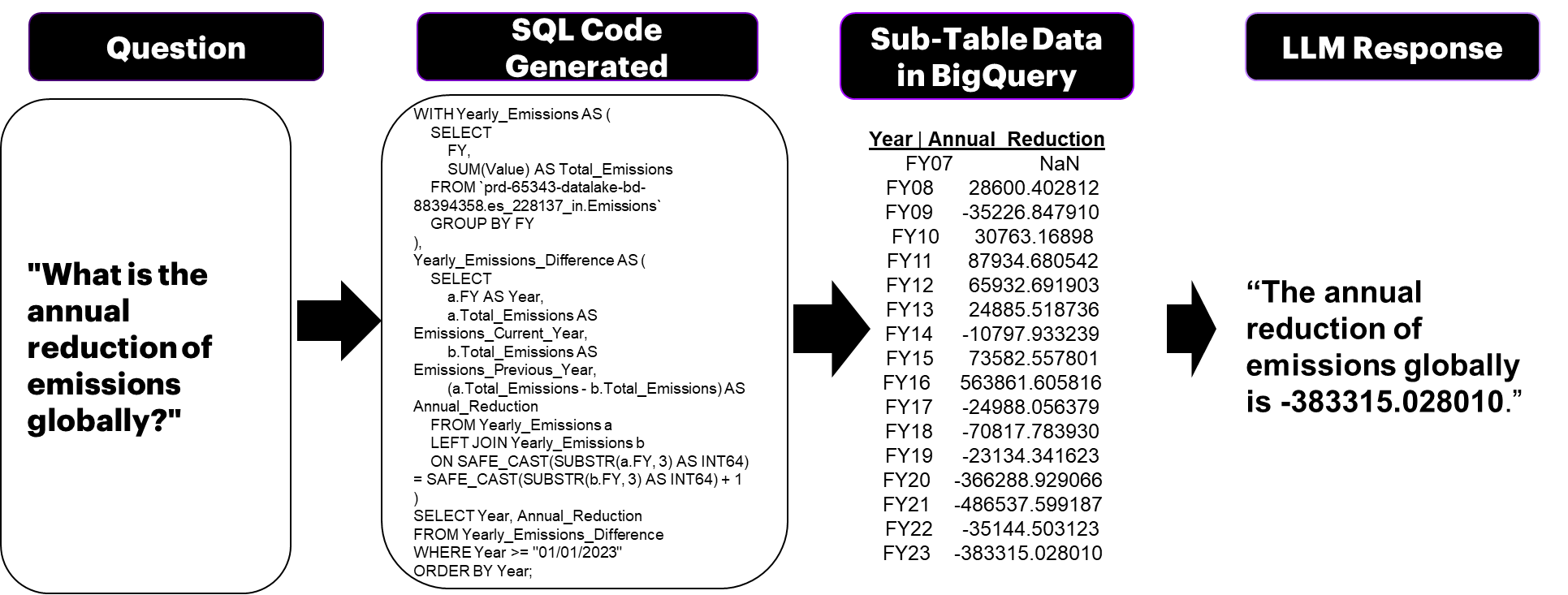}
    \caption{Example of the steps in Prompts 1, 2 and 3 to retrieve an appropriate response from the sustainability data tables.}
    \label{fig:res}\vspace{-0.3cm}
\end{figure*}

Finally, the third prompt per user-query utilizes the re-written and extended questions and the tabular data loaded at the end of prompt 2 in BigQuery to generate a customized prompt that is then sent to the LLM for a natural language response. This step is an extension to the classical \textit{seq2sql} models as it collects the SQL query outputs combines them with standard instructional guardrails and a sample question and answer for response generation. The style and format are specified as part of the prompt. Additional instructions are provided to display relevant error messages related to lack of tabular-data access, lack of data and irrelevant user-queries. It is noteworthy that per user-query the requirement is for all 3 prompts to complete under 10 seconds of run time to ensure real-time qualities. An example of the three prompts per user query is shown in Fig. \ref{fig:res}.

\subsection{Quality Assurance}
LLMs have been well known to be riddled with \textit{hallucinations} \cite{bigdata} even after setting model paremeters like temperature and top-P to zero. We propose the following factual checks (flags) $[s_{1,i},s_{2,i},s_{3,i},s_{4,i},s_{5,i}]$ for each user-query ($i$) in combination with the outcomes in prompts 2 and 3 to detect possible hallucinations for the extreme-RAG responses. For these checks, \textit{spacy}-based modules are used to extract and compare the named entities in the user-query ($Q_{Ni}$), with those in the response ($R_{Ni}$), in the SQL queries ($\Sigma_{Ni}$) and the data loaded in BigQuery by Prompt 2 ($D_{Ni}$) as follows.
\begin{enumerate}
    \item {\bf Number check}: This flag in prompt 3 ensures that all numerical values mentioned in the answers are directly derived from the underlying data sources without any discrepancies or fabrications by the end of Prompt 2. This check is crucial for maintaining the factual accuracy of numerical data presented in responses as shown in \eqref{s1}.
    \item {\bf Entity Check}: This flag verifies that all entities mentioned in the user-query are accurately reflected in the response, ensuring the response's relevance and completeness. This step is critical for confirming that the answer addresses all aspects of the query comprehensively, thereby enhancing user satisfaction and trust in the system. For example, in a user-query regarding ``water consumption rates in the USA'', this check ensures that the named entity ``USA'' is included in the response ($R_{Ni}$) and query ($Q_{Ni}$) as shown in (2).
    \item {\bf Query Check}: This flag confirms that all essential keywords and conditions stated in the user question are included in the SQL commands that are executed in Prompt 2. This step verifies the presence of specific filter words used in the user question ($Q_{Ni}$) within the SQL queries ($\Sigma_{Ni}$). This flag is configured using column metadata for each table and each filter check is added individually per data schema. This step ensures that the query accurately reflects the criteria and constraints specified in the question and can be envisioned by equation (2) wherein $R_{Ni}$ is changed to $\Sigma_{Ni}$.
    \item {\bf Regurgitation check}: This warning flag identifies if the response merely replicates information from the prompt 3 without paraphrasing. This step checks for word sequences where ten or more consecutive words from the prompt 3 are repeated. This step detects responses that fail to expand upon the query, which is essential for providing value-added insights to the user \cite{bigdata}.
    \item Increase/Decrease {\bf Modifier check}: This flag checks for consistency between the directional changes mentioned in the user query and that described in the prompt 3 response. This step ensures that the response accurately reflects the dynamics or trends requested by the user, critical for analyses involving changes over time or comparisons. 
\end{enumerate}
\begin{align}\label{s1}
s_{1,i}=\left\{
        \begin{array}{ll}
		1  & \mbox{if } numbers(R_i) \subset numbers(D_{i}) \\
		0 & otherwise
	\end{array}
 \right\}, \\ \nonumber
s_{2,i}=\left\{
        \begin{array}{ll}
        		1  & \mbox{if } Q_{N_i}==R_{N_i} \\
        		0 & otherwise \\
        \end{array}
\right\}.\\
 \end{align}
\section{Experiments and Results}
For validation of the proposed system we have deployed the proposed extreme-RAG system to answer specific questions across domains and databases using the Vertex AI platform in Google cloud platform. Prompt 1 and 3 are built using TextBision002 while Prompt 2 is built using the Unicorn LLM. The proposed system is scalable and can be generalized to other cloud-based and on-prem LLMs as well.
\subsection{Qualitative Scalability Analysis}
We assess the scalability of our extreme RAG-system on 100 random queries from multiple domains such as financial reporting, healthcare and social-media. Sample user-queries along with the generated SQL and the quality check scores are shown in Table \ref{queries}. Here, we observe that although the prompt 3 contains instructions based on the sustainability dataset, using a new-domain for validation dataset results in similar response performances. For financial domain data tables and queries, the proposed system achieves highest confidence scores (all scores are 1). For healthcare related queries, we observe some numerical hallucinations in the patient ID numbers that leads to inaccurate numerical predictions (80\% accuracy for $s_1$ metric). For the social-media use case, the proposed system hallucinates numbers (that can be flagged by our scoring module) and the modifier flag returns a not applicable value of -1.
\begin{table*}[ht] 
\caption{Sample Queries, generated SQL and Response Scores}
\scalebox{0.75}
    {
\begin{tabular}{|c|c|c|}
\hline
{\bf Question}&{\bf SQL command}& {\bf $[s_1,s_2,s_3,s_4,s_5]$}\\
\hline
What was the revenue growth for the Northeast region in&SELECT region, SUM(revenue) AS revenue FROM financials WHERE region $=$ `Northeast'& [1,1,1,1,1]\\ 
last quarter compared to the previous quarter?&AND (quarter $=$ `Last' OR quarter $=$ `Previous') GROUP BY quarter&\\ \hline
How many new patients were admitted to the oncology department & SELECT department, COUNT (patient-id) AS new-patients FROM hospital-admissions&[0.8,1,1,1,1]\\
this month and how does this compare to last month?& WHERE department $=$ `Oncology' AND (month $=$ `This' OR month $=$ `Last') GROUP BY month&\\ \hline
What was the click-through rate (CTR) for the digital marketing campaign& SELECT platform, AVG (click-through-rate) AS CTR FROM marketing-data&[0,1,1,1,-1]\\
on social media platforms in January?&WHERE campaign $=$ `Digital Marketing' AND platform $=$ `Social Media' AND month$=$ `January'&\\\hline
\end{tabular}\label{queries}
}
\end{table*}

\subsection{Quantitative Response Analysis}
\begin{figure}[ht!]
    \centering
    \includegraphics[width=3.2in, height=1.5in]{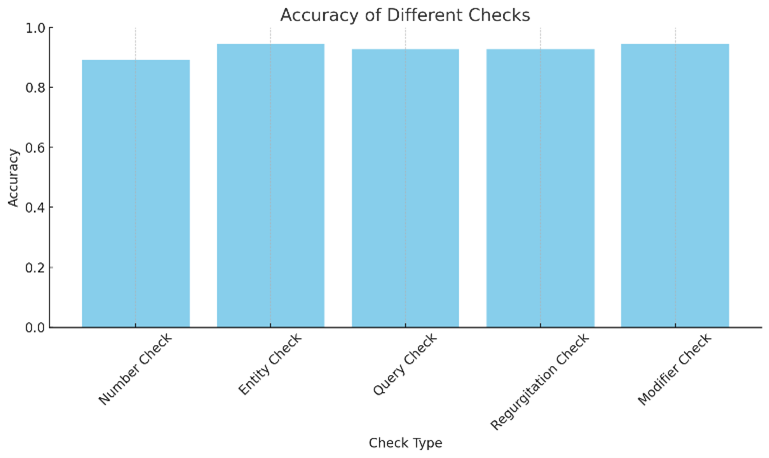}
    \caption{Averaged response quality scores on sustainability dataset [s$_1$ to s$_5$]}
    \label{fig:result}
\end{figure}

The quantitative responses on 500 sample variants of 60 queries generated on the sustainability datasets are shown in Fig. \ref{fig:result}. We observe that all responses have average scores around 90\%. The number check score is lowest at 89.09\% followed by entity check and modifier checks at 94.55\% each. Averaged query-check and regurgitation-check are around 92.73\%. This shows the impact of the proposed Extreme-RAG system towards hallucination control and scalability.
\subsection{Extreme-RAG Extensions}
\begin{figure}[ht!]
    \centering
    \includegraphics[width=3.2in]{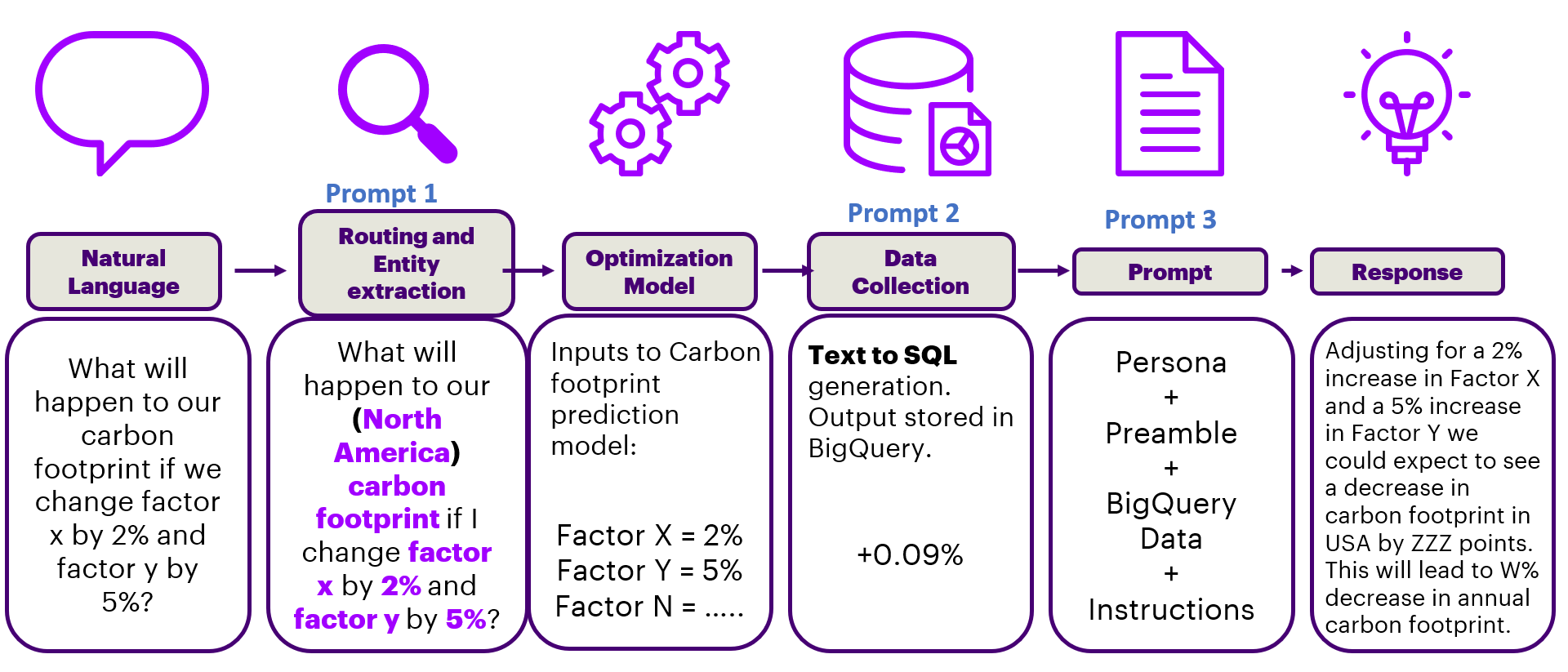}
    \caption{Example of an extension of the proposed Extreme RAG system to solve for predictive and prescriptive trend questions using back-end optimization algorithms that can be invoked by LLM prompts (prompt 2).}
    \label{fig:extend}
\end{figure}

The extreme-RAG architecture can further be extended to scenario planning and predictive capabilities using LLMs. Fig. \ref{fig:extend} shows a sample flow where a real-time optimization/predictive algorithm is invoked based on the named entities extracted from prompt 1, and the result returned from the algorithm can then be saved into BigQuery followed by a prompt 3 that furnishes the response in natural language. Such extensions/modifications to the code generation LLM prompt (prompt 2) enhance the scalability of the proposed system. 
\subsection{Comparison with Agentic-RAG}
Langchain (LangGraph) and LlamaIndex have been utilized heavily by several custom products till date to enable LLMs with orchestration capabilities to build intelligent AI agents \cite{llamaindex}. For any user-query where the underlying data is relatively large to skim through or highly varying in nature, AI agents can perform a web-search, knowledge identification followed by RAG and automated verification (using Trulens \cite{trulens}). In Table \ref{agent}, we compare the performance of an agentic-RAG incorporated using Langchain after prompt 1 to the proposed Extreme-RAG solution. It is noteworthy that agentic-RAG involves prior vector-embedding of the enterprise Tables as an offline process and an additional LLM-based evaluation step.
\begin{table}[ht]
\caption{Comparison with Agentic RAG after Prompt 1}
\scalebox{0.8}
    {
\begin{tabular}{|l|l|l|l|l|}
\hline
{\bf Method}&{\bf \# LLM calls}&{\bf avg time (s)}&{\bf Evaluation metrics}&Avg score\\\hline
Proposed&2&8.5&[$s_1$-$s_5$]&89-95\%\\
Langchain&4-5&40&[Relevance, Toxicity, Grounding]&40-98\%\\ \hline
\end{tabular}\label{agent}
\vspace{-0.3cm}
}
\end{table}

\section{Conclusion}\label {conclusion}
In this work, we propose a novel multi-LLM system that enables scalable tables-to-answers for Enterprise-level sustainability data tables with an averaged accuracy of $>90\%$ in under 10 seconds per user-query. The proposed process of dividing a user-query for query authentication, query understanding and routing, sub-tablular retrieval followed by answer generation is more stable and faster when compared to agentic-RAG approaches, which makes it a better fit for numeric-intensive use cases such as Sales, Financial and Sustainability domains. Future works can be directed towards extending RAG approaches to heterogeneous data sources for automation and insight generation tasks.
\bibliographystyle{IEEEtran}
\bibliography{main}

\end{document}